\documentclass[runningheads]{llncs}

\usepackage{eccv}

\usepackage{eccvabbrv}

\usepackage{graphicx}
\usepackage{booktabs}
\usepackage{multirow}
\usepackage{makecell}
\usepackage{hyperref}
\hypersetup{colorlinks=true, linkcolor=blue, urlcolor=blue, citecolor=blue}
\usepackage[accsupp]{axessibility}  

\usepackage{hyperref}

\usepackage{orcidlink}

\begin{document}

\title{Histopathology Multi-modal Embedding for Pathology Composed Retrieval} 

\titlerunning{HOMIE}

\author{Qifeng Zhou\inst{1} \and
Wenliang Zhong\inst{1} \and Thao M. Dang\inst{1} \and Hehuan Ma\inst{1} \and Saiyang Na\inst{1} \and Yuzhi Guo\inst{1} \and
Junzhou Huang\inst{1}}

\authorrunning{Q. Zhou et al.}

\institute{The University of Texas at Arlington, Arlington TX 76010, USA\\
\email{\{qxz8706, wxz9204, tmd4090, hehuan.ma, sxn3892, yuzhi.guo\}@mavs.uta.edu \\ jzhuang@uta.edu}}

\maketitle

\begin{abstract}
To overcome the black-box nature of predictive AI and the hallucination risks of generative models, retrieval-based models offer an interpretable, evidence-based paradigm for pathology clinical workflow. However, real-world clinical queries are inherently interleaved (e.g., pathology images and text). Current dual-encoders suffer from an \textbf{Architectural Mismatch}, lacking the mechanism to fuse such composed queries. To address this, we formalize the task of Pathology Composed Retrieval (PCR). While Multimodal Large Language Models (MLLMs) offer deep-fusion capabilities, directly applying them exposes a \textbf{Task Mismatch} and a \textbf{Domain Mismatch}. To resolve these challenges, we propose HOMIE, a model-agnostic adaptation framework that transforms any generative MLLM into a specialized pathology retrieval expert. Evaluated on our newly introduced PCR Benchmark, a lightweight 2B-parameter HOMIE variant substantially outperforms existing paradigms, surpassing specialized 7B pathology MLLMs and dual-encoders by large margins on composed retrieval, while maintaining strong performance on traditional simple retrieval. The project page is available at \url{https://qfchou.github.io/HOMIE_page/}.
\end{abstract}

\section{Introduction}
The digitization of pathology has created a promising opportunity for Artificial Intelligence (AI) to enhance clinical diagnosis~\cite{lu2024visual, xiang2025vision, sun2024pathmmu}. However, current AI models in this high-stakes domain face trust barriers. Traditional supervised models are ``black boxes'' that lack transparency, while emerging Large Language Models (LLMs) introduce the clinically unacceptable risk of hallucination~\cite{guidotti2018survey, huang2025survey}. Given these risks, a retrieval-based paradigm is more suitable for clinical application~\cite{bera2019artificial, chen2022fast}. Rather than predicting a diagnosis, a retrieval system functions as a ``computational consult''~\cite{shmatko2022artificial}. It searches its database for relevant information based on a pathologist's query~\cite{kalra2020yottixel, alfasly2025validation}. This approach empowers pathologists to examine this information and draw their own expert-guided conclusions, preserving clinical autonomy and fostering trust~\cite{cai2019human, hu2024histopathology}. Therefore, a powerful, specialized retrieval model is important for advancing scalable AI in pathology.

While recent multi-modal models have enabled basic pathology retrieval~\cite{lu2024visual,xiang2025vision}, they are limited to simple retrieval (e.g., image-to-text or text-to-image) by their dual-encoder architectures. Real-world clinical queries in pathology are inherently compositional and interleaved. Architecturally, dual-encoders lack a mechanism to deeply fuse such interleaved multi-modal inputs, resulting in an \textbf{Architectural Mismatch}. Forcing these models to perform composed retrieval may discard critical cross-modal semantic interactions. To address this gap, we formally define the task of \textbf{Pathology Composed Retrieval (PCR)}: retrieving the most relevant items from a database given a query composed of an interleaved sequence of pathology images, texts, and videos. This task requires a multi-modal embedding model, which can generate a single, semantically embedding from any combination of modalities for complex retrieval tasks.

A natural intuition to overcome this architectural mismatch is Multimodal Large Language Models (MLLMs), whose deep-fusion architectures natively process interleaved inputs~\cite{liu2024visual}. However, directly applying MLLMs to PCR presents some challenges. First, \textbf{Task Mismatch}: MLLM's latent spaces are optimized for generation, lacking the discriminative metric space required for retrieval~\cite{jiang2024e5}. Our empirical evaluations reveal that even state-of-the-art pathology MLLMs (e.g., PathoR1~\cite{zhang2025patho}) perform poorly on PCR tasks. Furthermore, general MLLMs adapted for retrieval suffer from \textbf{Domain Mismatch}, as they cannot interpret pathology features like subtle cellular morphology or staining artifacts~\cite{shen2022randstainna}. To solve these challenges, we propose HOMIE framework. Crucially, HOMIE is not a monolithic model, but a \textbf{model-agnostic, systematic adaptation framework} designed to transform any generative MLLM into a specialized pathology retrieval expert capable of generating a unified multi-modal embedding.  

HOMIE systematically resolves all three aforementioned mismatches. First, by leveraging an MLLM backbone, HOMIE inherently overcomes the \textbf{Architectural Mismatch}. To resolve the remaining barriers, we employ a two-stage process. A text-only retrieval-adaptation stage forces the generative LLM to learn a discriminative metric space, directly solving the \textbf{Task Mismatch}. Subsequently, a pathology-specific tuning stage addresses the \textbf{Domain Mismatch} by implementing dynamic native-resolution processing, targeted stain augmentation, and a progressive knowledge curriculum. To evaluate this, we also introduce the first PCR Benchmark, enabling quantification of these new tasks.

Our main contributions are summarized as follows: (1) We formally define the novel PCR task and introduce the PCR Benchmark for its rigorous evaluation. (2) We propose HOMIE, a novel, model-agnostic adaptation framework that resolves the Architectural, Task, and Domain mismatches, successfully transforming generative MLLMs into pathology retrieval experts. (3) Extensive experiments demonstrate that HOMIE outperforms specialized dual-encoder and MLLM-based methods by large margins on PCR tasks and maintains strong performance on traditional simple retrieval.

\section{Related Work}
\textbf{Vision-language Models in Pathology.}
Recent studies in pathology have adopted the CLIP~\cite{radford2021learning} model, leveraging paired image-text data within a contrastive learning framework to align similar image-text embeddings while separating dissimilar ones. These models, including PubmedCLIP~\cite{eslami2023pubmedclip}, BiomedCLIP~\cite{zhang2023biomedclip}, PLIP~\cite{huang2023visual}, PathCLIP~\cite{sun2024pathasst}, QuiltNet~\cite{ikezogwo2024quilt}, PathgenCLIP~\cite{sunpathgen}, PathoCLIP~\cite{zhang2025patho}, trained on amounts of pathology image-text pairs, excel at basic cross-modal retrieval tasks like image-to-text and text-to-image search. More advanced models, such as CONCH~\cite{lu2024visual} and MUSK~\cite{xiang2025vision}, achieved stronger performance by adopting more sophisticated architectures (CoCa~\cite{yu2022coca} for CONCH and BEiT3~\cite{wang2023image} for MUSK) and incorporating generative captioning objectives alongside contrastive alignment, surpassing the performance of simple CLIP-based methods. However, all these methods are constrained by an architectural paradigm reliant on separate encoders for image and text. The input is either a single text or a single image. Furthermore, all these models process images at a fixed, low resolution (e.g., 336x336), discarding the fine-grained morphological details critical for pathological analysis. Our framework is the first to move beyond this rigid design, using a unified MLLM to generate a true omni-modal embedding from interleaved, multi-modal data.

\noindent\textbf{Multimodal Embedding Learning.} MLLMs have extended LLMs to process multiple data modalities, achieving notable progress in understanding and reasoning across diverse input types~\cite{liu2024visual, liu2024improved, li2024llava}. To adapt MLLMs for retrieval tasks, the recent work E5-V~\cite{jiang2024e5} fine-tunes an LLM with summarization prompts and text-only data to extract embeddings, then integrates a vision module to obtain multimodal embeddings for zero-shot multimodal retrieval tasks. VLM2Vec~\cite{jiangvlm2vec}, LamRA~\cite{liu2024visual} and GME~\cite{zhang2024gme} leverage multimodal data and prompt-tuning to accommodate diverse queries and modalities, achieving strong performance on multimodal retrieval tasks. However, these models are designed for and pre-trained on general-domain data. This creates Domain Mismatch, as these models are not equipped to interpret pathology images, such as subtle morphological variations or staining artifacts. For pathological adaptation, directly applying these frameworks requires a domain-specific LLM or MLLM pretrained on pathology instruction data, which demands detailed and standardized annotations. In contrast, our framework, HOMIE can adapt an MLLM to generate omni-modalembeddings for pathology composed retrieval with only public pathology image-text pairs, bypassing these prohibitive requirements.

\section{Methods}
\label{sec:formatting}
\subsection{Pathology Composed Retrieval Benchmark}
Formally, we define the Pathology Composed Retrieval (PCR) task as follows: Given a query $q$ and a large candidate set $\mathcal{C} = \{c_1, c_2, \cdots, c_N\}$, the aim is to find similar candidates in $\mathcal{C}$ based on their similarity scores to $q$. A key feature of PCR is that both the query $q$ and each candidate $c_i$ are multi-modal, meaning they can be a single image, a single text, a single video, or any interleaved sequence.

To address the critical evaluation gap, we introduce the Pathology Composed Retrieval (PCR) Benchmark. This benchmark is designed to test a model's ability to perform compositional retrieval, moving beyond simple image-text retrieval. We created a suite of novel pathology composed retrieval tasks by repurposing existing public datasets:
\begin{table}[!t]
\small
\centering
\caption{The statistics of our PCR benchmark. It includes two retrieval tasks with selected samples from six data sources. Our benchmark has various combinations of Image (Img.), Text (Txt.), and Video (Vid.) modalities for query and target sides.}
\setlength{\tabcolsep}{4.5pt}
\scalebox{0.95}{
\begin{tabular}{llccc}
\toprule
Task & Data Source & Query Mod. & Target Mod. & $\#$ of Samples\\
\midrule
\multirow{5}{*}{\makecell[l]{Composed\\Retrieval}} 
& Bookset~\cite{gamper2021multiple} & Multi Img. & Txt. & 600\\
& Bookset~\cite{gamper2021multiple} & Img. \& Txt. & Img. & 488\\ 
& Quilt-VQA~\cite{seyfioglu2024quilt} & Img. \& Txt. & Txt. & 724\\ 
& Quilt-VQA-RED~\cite{seyfioglu2024quilt} & Img. \& Txt. & Txt. & 252\\ 
& Videopath~\cite{vuong2025videopath} & Vid. & Txt. & 244\\ 
\midrule
\multirow{6}{*}{\makecell[l]{Simple\\Retrieval}} 
& Bookset~\cite{gamper2021multiple} & Img. & Txt. & 2,688 \\ 
& Educontent~\cite{sun2024pathmmu} & Img. & Txt. & 947 \\ 
& MMUpubmed~\cite{sun2024pathmmu} & Img. & Txt. & 1,385 \\ 
& Bookset~\cite{gamper2021multiple} & Txt. & Img. & 2,688\\ 
& Educontent~\cite{sun2024pathmmu} & Txt. & Img. & 947\\ 
& MMUpubmed~\cite{sun2024pathmmu} & Txt. & Img. & 1,385\\ 
\bottomrule
\end{tabular}
}
\label{t1}
\end{table}

\noindent\textbf{Multi-Image to Text Retrieval $(q^i, q^i, ...) \to c^t$:} This task measures a model's ability from multiple images to retrieve a single, corresponding caption. We extract from Bookset~\cite{gamper2021multiple}, which provides multi-image with a single caption.

\noindent\textbf{Image-Text to Image Retrieval $(q^i, q^t) \to c^i$:} This task tests fine-grained compositional reasoning. Instead of generating new content, we utilize GPT-5 strictly to parse the original expert diagnostic texts from the multi-image set in Bookset~\cite{gamper2021multiple} and extract existing relational clauses (e.g., "a magnified view of the atypia..."). This creates a query (source image + relational text) to retrieve the correct target image while preserving absolute clinical fidelity.

\noindent\textbf{Image-Text to Text Retrieval $(q^i, q^t) \to c^t$:} This task reframes Visual Question Answering (VQA) to simulate pathologist inquiry and we extract from Quilt-VQA~\cite{seyfioglu2024quilt} and Quilt-VQA-RED~\cite{seyfioglu2024quilt}. To prevent models from ``cheating'' by exploiting lexical overlap between questions and answers, we apply GPT-5 purely as a linguistic formatter to restructure the phrasing of the answer. This ensures the underlying clinical semantics remain unaltered while forcing the model to genuinely ground its reasoning in the visual evidence.

\noindent\textbf{Video-to-Text Retrieval $q^v \to c^t$}: This task is extracted from VideoPath dataset~\cite{vuong2025videopath}. The model should retrieve a text diagnostic description with a video query, testing the fusion of spatio-temporal features. 

As shown in Table~\ref{t1}, these tasks can directly measure a model's ability to generate a true multi-modal embedding for pathology. Detailed data curation protocols and the exact LLM prompts, which are explicitly engineered to strictly prohibit the generation of new content, ensuring the rewriting process serves solely as a linguistic formatter while absolutely preserving the original medical facts and clinical fidelity, are provided in the supplementary material.

\subsection{Analysis of Existing Models on PCR}
Before detailing the specific architecture, we first analyze why existing models fail at PCR task. As shown in the bar charts of Figure~\ref{fig:moti} (bottom right), on the $(q^i, q^i, ...) \to c^t$ task, the best dual-encoder and MLLM-based model reach only 44.3\% and 14.7\% Recall@1, respectively, whereas HOMIE achieves 79.8\%. By investigating these empirical failures, we identify the fundamental mismatches that limit current models:
\begin{figure}[!t]
  \centering
\includegraphics[width=0.9\textwidth]{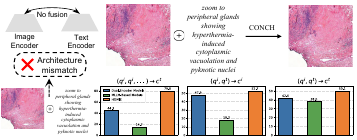} \\
  \caption{Analysis of Existing Models on PCR. \textbf{Left}: Dual-encoder models face an \textit{Architectural Mismatch}, forcing them to process interleaved queries separately. \textbf{Top Right}: Dual-encoder models fail in composed retrieval and degenerate into shallow matching, erroneously retrieving the original image. \textbf{Bottom Right}: Empirical results on the PCR Benchmark. Both paradigms exhibit substantial performance degradation: dual-encoders due to \textit{Architectural Mismatch}, and MLLMs due to \textit{Task and Domain Mismatches}.}
 \label{fig:moti}
\end{figure}

\noindent \textbf{The Architectural Mismatch of Dual-Encoder Models.} As illustrated in Figure~\ref{fig:moti} (left), dual-encoder models separate visual and textual inputs. Because these models lack a mechanism to fuse interleaved multi-modal queries, they rely only on independent, shallow similarity scores. As shown in Figure~\ref{fig:moti} (top right), instead of performing true multi-modal reasoning to find the correct diagnostic target, these models often reduce to visual matching, erroneously retrieving the original unannotated image itself.

\noindent \textbf{The Task and Domain Mismatches of MLLMs.} MLLMs inherently possess the deep-fusion architecture needed for interleaved inputs. However, directly deploying them exposes two flaws. First, a Task Mismatch: their latent spaces are optimized for token generation rather than contrastive metric learning, making their hidden states unsuited for retrieval. Second, a Domain Mismatch: general-domain MLLMs lack the morphological priors necessary to interpret complex pathology data.

\noindent\textbf{The HOMIE Solution.} To systematically overcome these diagnosed barriers, we propose the HOMIE framework. We detail the architecture design and the two-stage adaptation strategy in the subsequent sections.

\subsection{Architecture and Omni-modal Embedding}
\begin{figure}[!t]
  \centering  \includegraphics[width=1.0\textwidth]{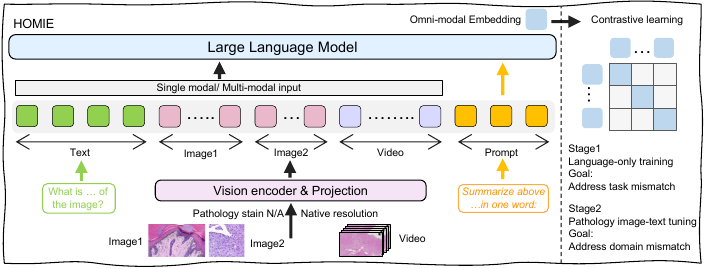} \\
  \caption{Overview of the HOMIE framework. The model ingests arbitrary modalities (e.g., image, text, video) and leverages a prompt-guided LLM to generate a unified omni-modal embedding. The vision pathway is optimized for pathology data, featuring pathology-specific stain normalization/augmentation and native resolution input. We employ a two-stage contrastive learning strategy to train the model, addressing task mismatch followed by domain mismatch.}
 \label{fig:arch}
\end{figure}

Our model consists of three critical components: a vision encoder $f_v$, an MLP-based projector $f_p$, and a LLM $f_{\varphi}$. The method overview is shown in Figure~\ref{fig:arch}. 

\noindent\textbf{Native Resolution and Video Processing.} To process visual inputs without discarding crucial diagnostic details, we utilize the SigLIP-2 ViT architecture~\cite{tschannen2025siglip}. For static images, instead of forcibly resizing inputs to a fixed, low resolution (e.g., $336 \times 336$), we dynamically adjust the image dimensions to multiples of a base patch size ($14 \times 14$). Then we apply 2D Rotary Positional Embeddings (2D-RoPE)~\cite{su2024roformer} to keep the position information. This design guarantees that the model processes pathology images at their native resolutions efficiently. For video inputs, we utilize frame rate sampling and interleaved multi-RoPE~\cite{huang2025revisiting} for robust spatio-temporal understanding.

\noindent\textbf{Multi-modal Embedding Extraction.} Given a visual input $V$ (image, video, or interleaved sequence), it is encoded and projected into a sequence of visual tokens $h_v=f_p(f_v(V))$. The LLM $f_{\varphi}$ then processes $h_v$ alongside any text tokens $h_t$ to produce a multi-modal embedding $E=f_{\varphi}(h_v, h_t)$. To extract a single, dense representation from a LLM, we employ the Explicit One-word Limitation (EOL) prompting strategy~\cite{jiang2024scaling}. The rationale behind EOL lies in the architectural difference between masked and causal language models. Unlike masked language models (e.g., BERT) that use bidirectional attention to inherently condense semantic information into a special \texttt{[CLS]} token, LLMs rely on causal attention. Conventional pooling strategies, such as mean pooling across all tokens, are sub-optimal for causal LLMs because early tokens lack access to future context, inherently diluting the global representation. Without explicit constraints, the hidden states of LLMs are naturally optimized for open-ended text generation rather than discriminative metric learning~\cite{jiang2024scaling}. Recent literature~\cite{jiang2024scaling} has empirically demonstrated that the EOL strategy overcomes this limitation and significantly outperforms traditional pooling methods, establishing it as the SOTA approach for extracting dense representations from LLMs. By explicitly appending the phrase ``in one word:'' to the prompt, we force the LLM to encapsulate and compress the entirety of the preceding multi-modal context into the target token's hidden state. Specifically, we use the following prompts: (i) for image-only input, we use: \texttt{<image> Summarize above image in one word: <emb>}; (ii) for text-only input, we use: \texttt{<text> Summarize above sentence in one word: <emb>}; (iii) for mixed image-text input, we utilize: \texttt{<image\textsubscript{1}><text\textsubscript{1}>...<image\textsubscript{i}><text\textsubscript{j}> Summarize above image and sentence in one word: <emb>.} where \texttt{<image>} and \texttt{<text>} denote placeholders for the input image and text, respectively. We use the last hidden state immediately preceding the \texttt{<emb>} token as the representation of the input.

\subsection{The HOMIE Framework}
We introduce HOMIE, a systematic framework to adapt an MLLM into an omni-modal, pathology-specific retrieval model. Our framework is a two-stage process. The first stage solves task mismatch by adapting LLM for retrieval using language-only pre-training. The second stage solves the domain mismatch by pathology-specific tuning. This framework is trained using only public data, which we systematically collated and curated to support this process.

\noindent\noindent\textbf{Training Dataset Curation.} Our training data includes text-only data and image-text pairs. Text-only data is used for Stage one. We use text pairs from Natural Language Inference (NLI) dataset~\cite{gao2021simcse}, MedNLI~\cite{romanov2018lessons} and MedMCQA~\cite{pal2022medmcqa}. To ensure high domain specificity, we curated a pathology-specific subset from MedMCQA~\cite{pal2022medmcqa}. Image-text pairs are used for Stage two and are aggregated from PathGen-1.6M~\cite{sunpathgen}, PathCap~\cite{sun2024pathasst}, and Quilt-1M~\cite{ikezogwo2024quilt}. We observed that naively mixing these heterogeneous sources leads to suboptimal performance. We identified the primary cause: Noisy image–text pairs collected from the web present challenges for training and may degrade the model performance. Therefore, a core part is bootstrap~\cite{li2022blip} approach: we first trained a base model on the unfiltered Quilt-1M data, then used this model to score the alignment of all pairs in Quilt-1M. All pairs with an image-text similarity score below a predefined threshold $\lambda=0.1$ were discarded. This curation and filtering process is the first step in our solution, ensuring that our specialization stages are built upon high-quality, domain-relevant data.

\noindent\textbf{Stage One: Adapting for Retrieval.} The first stage addresses task mismatch. MLLMs are optimized for generative tasks, not retrieval, meaning their latent space isn't structured for metric learning. To solve this issue, we train LoRA modules on our curated, pathology-specific text data to adapt the LLM for retrieval tasks. This stage forces the LLM to learn a semantically structured embedding space, making it suitable for retrieval before it ever sees images.

\noindent\textbf{Stage Two: Pathology-Specific Curriculum Tuning.} The second stage addresses the domain mismatch by fine-tuning the entire MLLM on our curated pathology image-text pairs. Different from natural images, pathology images present unique challenges, including chiefly staining variability and complex morphological features, which simple fine-tuning cannot solve. We therefore employ a pathology-specific tuning that incorporates three key adaptations. First, at the input level, we apply pathology stain augmentation and normalization~\cite{shen2022randstainna}, forcing the model to learn stain-invariant morphology. Second, we leverage our architecture's ability to process images at their native, original resolution. This is crucial as it ensures the model can analyze the multi-scale, fine-grained details that fixed-resolution models discard. Finally, we employ a Progressive Knowledge Curriculum instead of simply mixing datasets. This staged pathology curriculum tuning: the model is first exposed to Pathgen-1.6M~\cite{sunpathgen}, which emphasizes tissue-cell morphology and spatial organization to build foundational morphological priors. Then the model is trained on PathCap~\cite{sun2024pathasst} and our filtered Quilt-1M to learn how to associate these morphologies with high-level, multimodal diagnostic knowledge that includes diagnostic information.

\subsection{Training Objective}
We employ contrastive learning with the InfoNCE loss~\cite{radford2021learning} for both language-only pre-training and pathology tuning stages. 
Specifically, given a batch size of $B$, the embeddings of $i$-th query $q_i$ should be positioned close to the embeddings of its positive target $c_i$ and far away from other negative instances, formulated as: 
\begin{equation*}
\mathcal{L}=-\frac{1}{B} \sum_{i=1}^B \log \left[\frac{\exp \left[ cos\left(q_i, c_i\right) / \tau \right]}{\sum_{j=1}^B \exp \left[ cos\left(q_i, c_j\right) / \tau \right]}\right],
\end{equation*}
where $\tau$ is a temperature parameter and
$cos(q_i, c_i)$ represent the cosine similarity of $q_i$ and $c_i$ in contrastive learning.

\section{Experiments}
\textbf{Training Datasets and Metrics.} 
Our framework is trained entirely on publicly available data, as detailed in Sec 3.1. For stage one, we use the full NLI dataset~\cite{gao2021simcse} and a curated text corpus consisting of 14k pairs from MedNLI~\cite{romanov2018lessons} and 15k pathology-specific QA pairs filtered from MedMCQA~\cite{pal2022medmcqa}. For stage two, we follow our pathology progressive tuning. We first use 1.6M pairs from PathGen-1.6M~\cite{sunpathgen} to instill foundational knowledge in tissue-cell morphology and spatial organization. We then use 0.2M pairs from PathCap~\cite{sun2024pathasst} and 0.5M pairs from Quilt-1M~\cite{ikezogwo2024quilt} (after our rigorous filtering) to develop broader multimodal knowledge that includes diagnostic information. For the retrieval tasks, we primarily utilize Recall@K as the evaluation metric.

\noindent\textbf{Baseline Methods.} To demonstrate the effectiveness and necessity of HOMIE, we evaluate against two categories of baselines. We emphasize that existing paradigms fundamentally struggle with the PCR task due to architectural or objective mismatches. These comparisons aim to highlight the bottlenecks of current approaches rather than demonstrating the advantages of model scale. (1) \textbf{Pathology Dual-Encoder Models.} This group represents the current SOTA in pathology simple retrieval. We evaluate standard CLIP-based architectures, including PubmedCLIP~\cite{eslami2023pubmedclip}, BiomedCLIP~\cite{zhang2023biomedclip}, PLIP~\cite{huang2023visual}, PathCLIP~\cite{sun2024pathasst}, QuiltNet~\cite{ikezogwo2024quilt}, PathgenCLIP~\cite{sunpathgen}, PathoCLIP~\cite{zhang2025patho}, alongside advanced models like CONCH~\cite{lu2024visual} and MUSK~\cite{xiang2025vision} (based on CoCa~\cite{yu2022coca} and BEiT3~\cite{wang2023image}). Due to their rigid dual-encoder architectures, these models inherently lack the physical mechanism to deeply fuse interleaved multi-modal queries. To enable their evaluation on the PCR task, we employ parameter-free vector addition and reciprocal rank fusion (RRF), establishing the upper bound of traditional non-MLLM paradigms and exposing their \textbf{Architectural Mismatch}. (2) \textbf{MLLM-based Models.} We evaluate MLLMs by categorizing them into three sub-types. First, \textit{General MLLMs} (e.g., Qwen2.5-VL-7B~\cite{bai2025qwen2}, Qwen3-VL-2B~\cite{bai2025qwen3} and Qwen3-VL-8B~\cite{bai2025qwen3}) serve as our foundation control, lacking both retrieval optimization and pathology priors, thus suffering from both task and domain mismatches. Second, \textit{Retrieval-adapted General MLLMs} (e.g., E5-V~\cite{jiang2024e5}, LamRA~\cite{liu2025lamra}, and GME~\cite{zhang2024gme}) have been specifically instruction-tuned for cross-modal retrieval in natural images. Their catastrophic failure on the PCR benchmark strictly isolates and exposes the severe \textbf{Domain Mismatch}. Finally, \textit{Pathology MLLMs} (e.g., HuatuoGPT-Vision-7B~\cite{chen2024towards}, Lingshu-7B~\cite{xu2025lingshu}, and PathoR1-7B~\cite{zhang2025patho}) possess comparable parameter scales ($\sim$7B) and rich pathology priors. Their inclusion demonstrates that generative pathology knowledge does not naturally translate to a discriminative metric space, explicitly isolating the \textbf{Task Mismatch}. Together, these sub-types validate the absolute necessity of HOMIE's two-stage adaptation framework.
\begin{table}[!t]
\centering
\caption{Zero-shot composed retrieval performance, reporting Recall (R@k) metrics (in percentage \%) at various thresholds. Bold and underlined values indicate the best and second-best performance, respectively. Dashed lines (-) indicate non-applicability, as these methods cannot process video inputs. Add and RRF denote parameter-free vector addition and reciprocal rank fusion, respectively.}
\resizebox{\textwidth}{!}{
\begin{tabular}{l|ccc|ccc|ccc|ccc|ccc}
\hline
\multirow{2}{*}{Model} & \multicolumn{3}{c}{Bookset} & \multicolumn{3}{c}{Bookset} & \multicolumn{3}{c}{Quilt-VQA} & \multicolumn{3}{c}{Quilt-VQA-Red} & \multicolumn{3}{c}{Videopath}\\ 
& R@1 & R@5 & R@10 & R@1 & R@5 & R@10 & R@1 & R@5 & R@10 & R@1 & R@5 & R@10 & R@1 & R@5 & R@10\\
\hline 
\multicolumn{16}{c}{\textit{Pathology Dual-Encoder Models (Add)}} \\
\hline
PubmedCLIP~\cite{eslami2023pubmedclip} & 0.8 & 1.8 & 3.0 & 7.7 & 17.3 & 23.9 & 21.1 & 31.4 & 35.5 & 30.2 & 45.2 & 50.8 & - & - & - \\
BiomedCLIP~\cite{zhang2023biomedclip} & 19.0 & 40.0 & 57.2 & 17.8 & 39.9 & 52.6 & 10.9 & 24.0 & 31.1 & 19.0 & 33.7 & 46.0 & - & - & - \\
PLIP~\cite{huang2023visual} & 6.8 & 20.2 & 31.0 & 26.8 & 49.6 & 59.8 & 22.1 & 33.7 & 39.0 & 26.6 & 42.5 & 46.0 & - & - & - \\ 
PathCLIP~\cite{sun2024pathasst} & 11.2 & 36.8 & 48.5 & 22.4 & 44.6 & 57.9 & 21.3 & 36.3 & 40.6 & 33.7 & 46.4 & 50.8 & - & - & - \\ 
QuiltNet~\cite{ikezogwo2024quilt} & 12.3 & 28.2 & 37.7 & 26.6 & 45.5 & 52.6 & 18.9 & 29.4 & 34.7 & 28.2 & 36.9 & 41.3 & - & - & - \\
CONCH~\cite{lu2024visual} & 41.8 & 71.7 & 79.7 & 43.7 & 70.8 & 81.1 & 9.7 & 25.6 & 36.0 & 18.3 & 42.5 & 58.3 & - & - & - \\
Pathgen-CLIP-L~\cite{sunpathgen} & 22.5 & 47.7 & 57.5 & 39.8 & 70.8 & 80.7 & 26.4 & 39.6 & 44.8 & 36.9 & 49.2 & 53.2 & - & - & - \\
Pathgen-CLIP~\cite{sunpathgen} & 23.0 & 52.8 & 64.3 & 39.2 & 66.5 & 78.9 & 28.2 & 41.6 & 46.0 & 35.3 & 50.0 & 56.3 & - & - & - \\
MUSK~\cite{xiang2025vision} & 43.0 & 72.7 & 83.0 & 47.9 & 74.2 & 84.3 & 34.1 & 53.0 & 58.4 & 50.0 & 69.8 & 77.8 & - & - & - \\
Patho-CLIP-L~\cite{zhang2025patho} & 44.3 & 77.5 & 88.0 & 40.7 & 69.7 & 78.8 & 22.1 & 40.5 & 48.6 & 35.3 & 58.7 & 66.3 & - & - & - \\
Patho-CLIP~\cite{zhang2025patho} & 43.3 & 75.8 & 85.7 & 35.6 & 61.8 & 73.6 & 14.1 & 28.7 & 35.5 & 21.8 & 40.1 & 47.2 & - & - & - \\
\hline
\multicolumn{16}{c}{\textit{Pathology Dual-Encoder Models (RRF)}} \\
\hline
PubmedCLIP~\cite{eslami2023pubmedclip} & 0.4 & 1.7 & 3.0 & 2.5 & 9.8 & 16.4 & 1.7 & 7.5 & 14.4 & 3.6 & 15.5 & 27.8 & - & - & - \\
BiomedCLIP~\cite{zhang2023biomedclip} & 15.8 & 38.3 & 51.5 & 15.8 & 37.5 & 52.5 & 3.9 & 10.1 & 16.6 & 7.9 & 21.8 & 36.1 & - & - & - \\
PLIP~\cite{huang2023visual} & 5.0 & 17.8 & 23.5 & 12.1 & 35.2 & 53.9 & 2.9 & 10.5 & 15.1 & 8.3 & 24.2 & 35.3 & - & - & - \\ 
PathCLIP~\cite{sun2024pathasst} & 9.8 & 27.0 & 42.2 & 19.3 & 39.8 & 56.4 & 3.3 & 9.9 & 16.3 & 12.3 & 28.6 & 39.3 & - & - & - \\ 
QuiltNet~\cite{ikezogwo2024quilt} & 10.3 & 29.5 & 41.0 & 15.4 & 34.6 & 46.3 & 4.8 & 11.9 & 17.5 & 14.7 & 27.8 & 36.9 & - & - & - \\
CONCH~\cite{lu2024visual} & 36.5 & 68.5 & 78.3 & 34.4 & 63.9 & 75.6 & 4.3 & 13.5 & 23.6 & 13.5 & 33.3 & 49.2 & - & - & - \\
Pathgen-CLIP-L~\cite{sunpathgen} & 22.2 & 56.7 & 71.8 & 27.7 & 58.2 & 74.6 & 8.0 & 19.5 & 26.2 & 26.6 & 45.6 & 53.6 & - & - & - \\
Pathgen-CLIP~\cite{sunpathgen} & 19.7 & 51.3 & 66.5 & 28.3 & 59.8 & 71.5 & 9.9 & 19.9 & 27.5 & 23.0 & 44.0 & 54.0 & - & - & - \\
MUSK~\cite{xiang2025vision} & 44.3 & 80.3 & 88.3 & 34.0 & 68.9 & 82.2 & 17.5 & 30.5 & 39.5 & 37.3 & 62.3 & 71.4 & - & - & - \\
Patho-CLIP-L~\cite{zhang2025patho} & 43.2 & 76.8 & 86.5 & 33.6 & 62.3 & 77.0 & 9.8 & 22.4 & 30.1 & 25.0 & 48.8 & 59.1 & - & - & - \\
Patho-CLIP~\cite{zhang2025patho} & 42.8 & 74.3 & 85.8 & 30.1 & 58.2 & 71.1 & 6.9 & 16.2 & 22.8 & 14.7 & 33.7 & 44.8 & - & - & - \\
\hline
\multicolumn{16}{c}{\textit{MLLM-Based Models}} \\
\hline
Qwen2.5-VL-7B~\cite{bai2025qwen2} & 3.5 & 8.7 & 16.0 & 3.9 & 12.7 & 23.2 & 14.4 & 27.6 & 35.1 & 24.2 & 42.1 & 50.0 & 1.6 & 4.5 & 8.6 \\
Qwen3-VL-2B~\cite{bai2025qwen3} & 0.5 & 2.5 & 4.0 & 1.8 & 5.5 & 8.4 & 3.6 & 8.7 & 11.9 & 2.0 & 9.6 & 13.9 & 0.8 & 2.9 & 4.9 \\
Qwen3-VL-8B~\cite{bai2025qwen3} & 1.7 & 6.0 & 10.3 & 3.3 & 10.9 & 15.6 & 7.2 & 14.1 & 18.2 & 16.7 & 29.4 & 34.1 & 1.2 & 3.3 & 5.3 \\
E5-V-7B~\cite{jiang2024e5} & 2.3 & 8.8 & 12.5 & 2.0 & 6.4 & 10.9 & 3.6 & 7.6 & 10.4 & 9.1 & 15.5 & 20.6 & - & - & - \\
LamRA-7B~\cite{liu2025lamra} & 5.8 & 15.7 & 24.0 & 7.4 & 20.5 & 29.3 & 23.6 & 37.4 & 44.3 & 52.4 & 66.7 & 73.4 & 2.5 & 10.2 & 16.8 \\
GME-7B~\cite{zhang2024gme} & 4.3 & 14.3 & 22.2 & 3.7 & 14.5 & 22.1 & 23.5 & 33.3 & 39.9 & 48.8 & 61.1 & 67.1 & 5.3 & 10.7 & 17.2 \\
HuatuoGPT-V-7B~\cite{chen2024towards} & 3.5 & 11.3 & 18.8 & 6.1 & 18.9 & 25.8 & 11.6 & 22.5 & 29.0 & 19.0 & 34.1 & 40.9 & 4.9 & 14.8 & 22.5 \\
Lingshu-7B~\cite{xu2025lingshu} & 6.8 & 22.8 & 33.2 & 10.2 & 24.8 & 35.0 & 17.5 & 30.7 & 37.0 & 23.8 & 39.7 & 48.4 & 4.5 & 12.3 & 16.4 \\
PathoR1-7B~\cite{zhang2025patho} & 14.7 & 33.2 & 45.0 & 18.2 & 43.2 & 52.3 & 14.2 & 28.7 & 35.1 & 20.2 & 41.7 & 52.4 & 19.3 & 50.4 & 61.5 \\
\hline
HOMIE (Qwen3-VL-2B) & \underline{75.2} & \underline{94.2} & \underline{96.7} & \underline{49.8} & \underline{76.0} & \underline{85.9} & \underline{34.4} & \underline{61.9} & \underline{69.8} & \textbf{64.3} & \textbf{89.7} & \textbf{92.9} & \underline{22.1} & \underline{59.0} & \underline{76.6} \\
HOMIE (Qwen3-VL-8B) & \textbf{79.8} & \textbf{96.7} & \textbf{98.3} & \textbf{52.7} & \textbf{79.5} & \textbf{88.5} & \textbf{35.8} & \textbf{64.1} & \textbf{72.8} & \underline{57.5} & \underline{89.3} & \textbf{92.9} & \textbf{30.7} & \textbf{65.6} & \textbf{79.9} \\
\hline
\end{tabular}}
\label{t2}
\end{table}

\noindent\textbf{Implementation Details.} 
We implement HOMIE using PyTorch and DeepSpeed ZeRO-2. All training is performed on 8 NVIDIA H100 GPUs with BF16 precision and FlashAttention-2~\cite{dao2023flashattention} enabled. For stage 1, we train for 2 epochs with a global batch size of 576. The learning rate is set to $2 \times 10^{-4}$ with a cosine decay and 0.03 warmup ratio. In this stage, the vision encoder and projector are frozen. We apply LoRA to the LLM with $r=64$ and $\alpha=128$. For stage 2, we initialize from the Stage 1 checkpoint and train for 2 epochs with a global batch size of 384. The learning rate is adjusted to $1 \times 10^{-4}$. Crucially, we unfreeze the vision encoder and the projector to capture domain-specific features. The LLM is trained via LoRA with an increased rank of $r=128$ and $\alpha=256$.

\subsection{Zero-shot Composed Retrieval}
Table~\ref{t2} reports zero-shot composed retrieval results, closely mirroring real-world clinical workflows (e.g., interleaved images and text). These results validate our framework while exposing the limitations of existing paradigms.

\noindent\textbf{Architectural Mismatch of Dual-Encoders.} Lacking internal mechanisms for multi-modal fusion, SOTA dual-encoders (e.g., MUSK~\cite{xiang2025vision}, CONCH~\cite{lu2024visual}) are restricted to parameter-free late-fusion heuristics, such as vector addition and reciprocal rank fusion (RRF). Consequently, their performance saturates at $\sim$44\% Recall@1 on the Multi-Image to Text task.

\noindent\textbf{Domain and Task Mismatches in MLLMs.} Directly deploying generative MLLMs is insufficient. First, general-domain MLLM-based methods perform poorly, reflecting a Domain Mismatch with complex pathology morphology. Second, despite possessing rich medical priors, $\sim$7B pathology MLLMs reach only 14.7\% R@1. This indicates a Task Mismatch.

\noindent\textbf{The Efficacy and Efficiency of HOMIE.} By contrast, HOMIE addresses these bottlenecks. Our lightweight HOMIE (Qwen3-VL-2B) variant achieves 75.2\% R@1, outperforming the best dual-encoders and exceeding PathoR1-7B~\cite{zhang2025patho} by over 60 percentage points. This suggests that HOMIE's gains stem from our two-stage adaptation rather than parameter scaling. Further scaling to an 8B backbone shows that HOMIE scales as a model-agnostic paradigm for composed pathology retrieval.

\subsection{Zero-shot Simple Retrieval}
\begin{table}[!t]
\small
\centering
\caption{Performance comparison with baseline methods on simple retrieval datasets, reporting Recall (R@k) metrics (in percentage \%) at different thresholds. Slash-separated values indicate image-to-text (i2t) / text-to-image (t2i) retrieval performance, respectively. Bold and underlined values indicate the best and second-best performance, respectively.}
\resizebox{\textwidth}{!}{
\begin{tabular}{l|ccc|ccc|ccc}
\hline
\multirow{2}{*}{Model} & \multicolumn{3}{c}{Bookset} & \multicolumn{3}{c}{Educontent} &\multicolumn{3}{c}{MMUpubmed} \\ 
& R@1 & R@5 & R@10 & R@1 & R@5 & R@10 & R@1 & R@5 & R@10\\
\hline
\multicolumn{10}{c}{\textit{Pathology Dual-Encoder Models}} \\
\hline
PubmedCLIP~\cite{eslami2023pubmedclip} & 0.2/0.1 & 1.0/0.8 & 1.6/1.2 & 0.2/0.4 & 1.3/1.6 & 2.6/2.6 & 0.4/0.1 & 1.5/0.8 & 3.0/1.2 \\
BiomedCLIP~\cite{zhang2023biomedclip} & 9.9/9.4 & 23.5/23.4 & 32.0/32.0 & 3.4/4.0 & 8.4/9.1 & 14.9/15.5 & 6.2/6.6 & 16.5/19.2 & 23.4/26.1 \\
PLIP~\cite{huang2023visual} & 3.3/2.3 & 9.2/8.1 & 14.2/13.2 & 1.2/2.0 & 6.2/6.3 & 9.0/9.7 & 0.7/1.4 & 4.4/5.2 & 7.7/8.8 \\ 
PathCLIP~\cite{sun2024pathasst} & 9.3/8.8 & 22.7/21.3 & 30.7/29.5 & 3.8/4.3 & 11.8/11.1 & 16.9/16.5 & 6.4/8.3 & 16.0/18.8 & 23.2/26.1 \\ 
QuiltNet~\cite{ikezogwo2024quilt} & 3.6/2.9 & 12.6/9.5 & 18.2/15.6 & 3.1/3.7 & 8.1/8.9 & 13.4/13.0 & 2.1/2.2 & 6.4/6.5 & 10.8/10.0 \\ 
CONCH~\cite{lu2024visual} & 25.0/23.6 & 50.0/48.7 & 62.2/59.9 & 5.4/6.2 & 15.8/15.9 & 22.4/23.5 & 6.1/7.1 & 17.4/19.1 & 25.0/26.5 \\
Pathgen-CLIP-L~\cite{sunpathgen} & 12.8/14.7 & 28.7/33.9 & 37.2/44.7 & 5.8/8.6 & 16.9/22.1 & 24.1/31.4 & 6.4/8.4 & 18.4/20.3 & 27.1/29.0 \\
Pathgen-CLIP~\cite{sunpathgen} & 15.0/14.1 & 32.3/34.2 & 41.4/45.2 & 7.4/9.0 & 18.9/21.6 & 26.5/32.1 & 4.5/6.3 & 14.1/15.5 & 21.7/23.7 \\
MUSK~\cite{xiang2025vision} & 23.9/25.3 & 49.0/50.1 & 62.3/62.0 & \underline{11.1}/9.0 & 26.6/28.2 & 35.2/38.2 & 10.6/\underline{11.0} & 25.6/26.7 & 35.8/33.5 \\
Patho-CLIP-L~\cite{zhang2025patho} & 22.5/21.9 & 50.6/49.3 & 62.0/62.6 & 4.2/3.6 & 13.4/13.4 & 21.8/21.0 & 4.0/3.8 & 11.9/10.3 & 17.8/16.2 \\
Patho-CLIP~\cite{zhang2025patho} & 26.7/27.6 & 55.1/55.8 & 67.2/67.6 & 3.7/4.0 & 10.9/12.4 & 16.5/18.0 & 2.7/3.2 & 9.2/9.2 & 14.9/13.4 \\
\hline
\multicolumn{10}{c}{\textit{MLLM-Based Models}} \\
\hline
Qwen2.5-VL-7B~\cite{bai2025qwen2} & 0.9/0.4 & 2.8/1.6 & 4.7/2.6 & 1.9/1.0 & 5.8/3.0 & 8.8/4.5 & 1.1/0.4 & 4.5/1.3 & 7.4/3.0 \\
Qwen3-VL-2B~\cite{bai2025qwen3} & 0.5/0.1 & 1.7/0.8 & 2.8/1.3 & 0.7/0.2 & 2.3/1.8 & 3.7/2.7 & 0.5/0.4 & 1.7/1.4 & 3.0/2.3 \\
Qwen3-VL-8B~\cite{bai2025qwen3} & 0.9/0.4 & 3.1/1.5 & 4.3/3.0 & 1.4/0.8 & 5.3/2.9 & 8.0/5.5 & 1.1/0.3 & 3.5/2.0 & 5.8/3.5 \\
E5-V-7B~\cite{jiang2024e5} & 1.7/0.5 & 5.5/1.8 & 8.6/2.4 & 2.6/0.7 & 7.7/1.6 & 9.7/3.2 & 2.3/0.8 & 7.8/2.6 & 10.7/4.5 \\
LamRA-7B~\cite{liu2025lamra} & 0.6/0.3 & 2.1/1.2 & 3.4/1.7 & 2.4/0.7 & 6.2/2.2 & 9.0/3.6 & 2.3/0.8 & 6.4/2.5 & 9.2/4.0 \\
GME-7B~\cite{zhang2024gme} & 1.9/0.5 & 5.1/2.3 & 7.8/4.1 & 4.1/1.5 & 8.8/3.7 & 11.7/5.5 & 3.3/1.4 & 7.4/3.8 & 10.5/6.0 \\
HuatuoGPT-V-7B~\cite{chen2024towards} & 0.9/0.4 & 4.6/2.0 & 7.0/3.2 & 1.9/1.2 & 5.6/3.4 & 8.7/4.5 & 1.2/0.3 & 5.3/2.7 & 8.5/4.0 \\
Lingshu-7B~\cite{xu2025lingshu} & 2.9/1.3 & 8.6/4.8 & 13.2/6.8 & 2.4/1.3 & 8.3/5.2 & 11.6/8.1 & 2.1/1.6 & 8.0/5.5 & 13.9/9.6 \\
PathoR1-7B~\cite{zhang2025patho} & 9.5/7.0 & 23.8/18.4 & 32.3/25.2 & 3.0/6.1 & 9.3/17.2 & 13.1/21.6 & 1.7/3.7 & 3.7/11.0 & 5.4/15.2 \\
\hline
HOMIE (Qwen3-VL-2B) & \underline{32.0}/\underline{29.7} & \underline{58.5}/\underline{54.7} & \underline{69.4}/\underline{65.1} & \underline{11.1}/\underline{12.7} & \underline{28.5}/\underline{29.7} & \underline{36.5}/\underline{38.3} & \textbf{13.3}/\underline{11.0} & \textbf{31.3}/\underline{27.1} & \underline{40.1}/\underline{35.5} \\
HOMIE (Qwen3-VL-8B) & \textbf{34.4}/\textbf{31.5} & \textbf{61.4}/\textbf{60.0} & \textbf{73.0}/\textbf{72.7} & \textbf{12.1}/\textbf{14.6} & \textbf{29.8}/\textbf{31.5} & \textbf{38.0}/\textbf{40.5} & \underline{13.1}/\textbf{11.1} & \underline{30.9}/\textbf{28.7} & \textbf{41.4}/\textbf{37.5} \\
\hline
\end{tabular}}
\label{t3}
\end{table}
A key question is whether gaining complex compositional reasoning capabilities compromises a model's performance on traditional simple (image-to-text and text-to-image) retrieval tasks. As shown in Table~\ref{t3}, HOMIE not only retains this fundamental capability but establishes a new SOTA. Both our 2B and 8B variants consistently outperform all specialized pathology dual-encoders (e.g., MUSK~\cite{xiang2025vision}, Patho-CLIP~\cite{zhang2025patho}) across the Bookset~\cite{gamper2021multiple}, Educontent~\cite{sun2024pathmmu}, and MMUpubmed~\cite{sun2024pathmmu} benchmarks. Furthermore, while other MLLM-based baselines continue to struggle due to inherent task and domain mismatches, HOMIE consistently outperforms them. This indicates that our two-stage adaptation framework instills fine-grained pathology knowledge into the MLLM. HOMIE achieves this performance using only public training data, without proprietary datasets.

\subsection{Generality of the HOMIE Framework}
A defining characteristic of HOMIE is that it is not a monolithic model, but rather a model-agnostic, systematic adaptation framework. To assess its generality, we swapped the core MLLM backbone with various foundation models, including general-domain models (Qwen series) and SOTA pathology-specific MLLMs. As shown in Table~\ref{t4}, HOMIE transforms every evaluated generative MLLM into a capable pathology retrieval model. Two observations emerge from these results. First, \textbf{synergy with domain priors}: while pathology-specific MLLMs (like PathoR1-7B) perform poorly at zero-shot retrieval on their own (as shown in Table~\ref{t2}), applying the HOMIE framework leverages their pathology knowledge. Consequently, HOMIE (PathoR1-7B) consistently outperforms HOMIE built on the vanilla Qwen2.5-VL-7B across almost all composed retrieval metrics. Second, \textbf{framework scalability}: by comparing Qwen3-VL-2B with Qwen3-VL-8B, we observe a consistent scaling trend. As the parameter size and reasoning capabilities of the base model increase, the retrieval performance of the HOMIE-adapted model increases accordingly. These results establish HOMIE as an extensible framework. This suggests that as more capable generative MLLMs become available, the framework offers a direct pathway to adapt them for pathology retrieval.
\begin{table}[!t]
\centering
\caption{Performance of the HOMIE framework across diverse MLLM backbones. We report Recall@1 (\%) metrics. Quilt-V. is abbreviated for Quilt-VQA.}
\renewcommand{\arraystretch}{1.25} 
\resizebox{\textwidth}{!}{
\begin{tabular}{l c c cc c ccc}
\toprule
\multirow{2}{*}{Model} & $(q^i, q^i, ...) \to c^t$ & $(q^i, q^t) \to c^i$ & \multicolumn{2}{c}{$(q^i, q^t) \to c^t$} & $q^v \to c^t$ & \multicolumn{3}{c}{$q^i \to c^t$}\\ 
\cmidrule(lr){2-2} \cmidrule(lr){3-3} \cmidrule(lr){4-5} \cmidrule(lr){6-6} \cmidrule(lr){7-9}
& Bookset & Bookset & Quilt-V. & Quilt-V.-Red & Videopath & Bookset & Educontent & MMUpubmed\\
\midrule 
HOMIE (Qwen3-VL-2B) & 75.2 & 49.8 & 33.4 & 64.3 & 22.1 & 32.0 & 11.1 & 13.3\\
HOMIE (Qwen3-VL-8B) & 79.8 & 52.7 & 35.8 & 57.5 & 30.7 & 34.4 & 12.1 & 13.1\\
HOMIE (Qwen2.5-VL-7B) & 77.2 & 49.2 & 32.5 & 62.3 & 28.3 & 30.0 & 12.2 & 11.7\\
HOMIE (HuatuoGPT-V-7B) & 76.3 & 47.5 & 32.0 & 57.9 & 28.7 & 31.3 & 11.6 & 12.1 \\
HOMIE (Lingshu-7B) & 78.0 & 50.6 & 33.8 & 59.9 & 26.6 & 31.4 & 12.6 & 13.4\\
HOMIE (PathoR1-7B) & 78.5 & 54.1 & 38.4 & 64.3 & 30.7 & 33.6 & 14.7 & 11.9 \\
\bottomrule
\end{tabular}
}
\label{t4}
\end{table}

\subsection{Qualitative Results and Embedding Space Visualization}
\begin{figure}[!t]
  \centering
  \includegraphics[width=\linewidth]{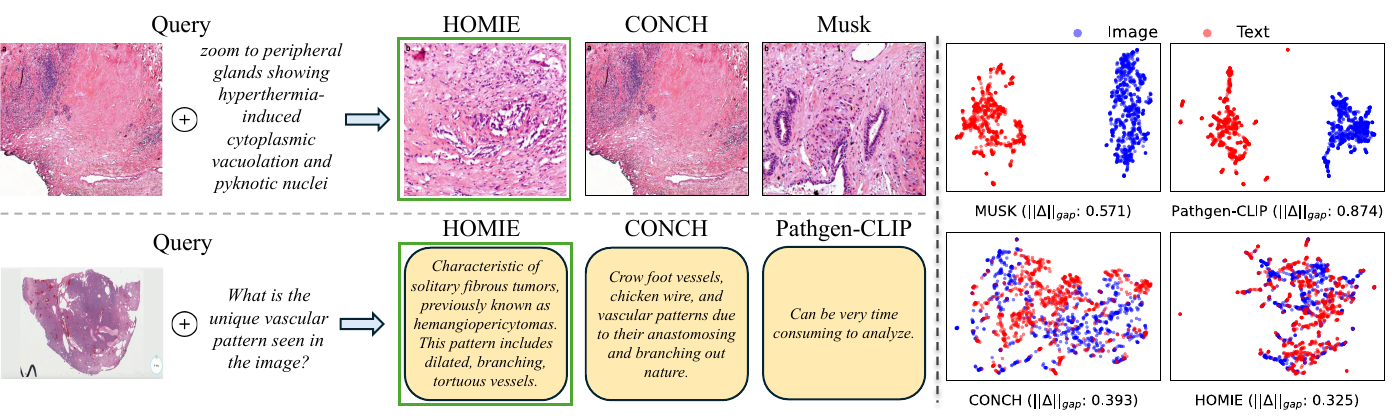}
  \caption{Qualitative results and visualization. \textbf{Left}: Examples of composed retrieval. HOMIE performs multi-modal reasoning, whereas dual-encoder models do not. \textbf{Right}: UMAP visualization of image (red) and text (blue) embeddings from the EduContent dataset. $||\Delta||_{gap}$ denotes the modality gap metric~\cite{liang2022mind}.}
   \label{f3}
\end{figure}

Figure~\ref{f3} (left) illustrates the practical consequences of the Architectural Mismatch. In the top Image-Text to Image example, HOMIE performs the required spatial-semantic reasoning. Conversely, CONCH~\cite{lu2024visual} ignores the textual instruction and erroneously retrieves the original source image, consistent with dual-encoders reducing to visual matching given interleaved inputs. Similarly, HOMIE grounds visual questions to diagnostic texts in the bottom example, whereas baselines return irrelevant descriptions. This behavior stems from our structurally aligned latent space. Figure~\ref{f3} (right) visualizes the modality gap via UMAP. While pathology CLIP baselines (e.g., Pathgen-CLIP~\cite{sunpathgen}, MUSK~\cite{xiang2025vision}) show a large modality gap, segregating features into separate clusters (high $||\Delta||_{gap} \ge 0.571$), HOMIE reduces this gap. By leveraging our two-stage adaptation, HOMIE achieves the lowest modality gap ($||\Delta||_{gap}=0.325$), mapping visual and textual concepts into a shared metric space rather than adjacent regions.

\subsection{Ablation Studies}
We conduct an ablation study to assess the contribution of each component, rather than attributing performance to scaling or curated data alone. As shown in Table~\ref{t5}, we disable each pathology-specific adaptation component on the HOMIE (Qwen3-VL-8B) backbone and observe consistent performance degradation across task categories.

\noindent\textbf{(1) Progressive Knowledge Curriculum:} Disabling the staged training causes the largest performance drop on complex reasoning tasks. For instance, Video-to-Text retrieval drops from 30.7\% to 18.4\%, and Image-Text to Image retrieval drops from 52.7\% to 46.7\%. This supports our hypothesis: an MLLM must first build foundational morphological priors before learning high-level, multi-modal diagnostic reasoning.

\noindent\textbf{(2) Native Resolution:} Forcing the model to use standard, low-resolution inputs leads to notable drops on tasks requiring fine-grained visual details. For example, Image-to-Text retrieval on Bookset~\cite{gamper2021multiple} drops by 8.9\%. This indicates that fine-grained morphological features discarded by fixed-resolution models are important for pathology retrieval.

\noindent\textbf{(3) Data Filtering:} Training on unfiltered, noisy web data degrades performance (e.g., from 30.7\% to 23.4\% on Video-to-Text). This indicates that our bootstrap filtering helps construct a well-aligned embedding space.

\noindent\textbf{(4) Stain Augmentation:} Removing pathology-specific stain normalization and augmentation leads to consistent performance drops across multiple tasks (e.g., a 2.1\% drop in Multi-Image to Text retrieval). This suggests it helps the model learn stain-invariant morphological representations. Together, these ablations indicate that each component contributes to the overall performance.

\noindent \textbf{(5) Prompting Robustness:} To understand the embedding extraction mechanism, we ablate our Explicit One-word Limitation (EOL) prompt at inference time. Extracting the last hidden state without explicit instructional constraints (\textit{No Prompt}) or with a weaker prompt (\textit{No One Word}) still yields competitive results. This robustness suggests that training adapts the model to condense multi-modal semantics into the final token, reducing reliance on the inference-time prompt. Applying the EOL constraint (\textit{HOMIE}) further aligns the inference behavior with the trained representation. The phrase ``in one word'' acts as a semantic bottleneck that reduces residual generation noise and yields the best performance (79.8\%).

\begin{table}[!t]
\centering
\caption{Ablation study of HOMIE's key components and prompts. We report the Recall@1 (\%) metrics. All evaluated models are based on the Qwen3-VL-8B backbone. Quilt-V. is abbreviated for Quilt-VQA.}
\renewcommand{\arraystretch}{1.25} 
\resizebox{\textwidth}{!}{
\begin{tabular}{l c c cc c ccc}
\toprule
\multirow{2}{*}{Ablations} & $(q^i, q^i, ...) \to c^t$ & $(q^i, q^t) \to c^i$ & \multicolumn{2}{c}{$(q^i, q^t) \to c^t$} & $q^v \to c^t$ & \multicolumn{3}{c}{$q^i \to c^t$}\\
\cmidrule(lr){2-2} \cmidrule(lr){3-3} \cmidrule(lr){4-5} \cmidrule(lr){6-6} \cmidrule(lr){7-9}
& Bookset & Bookset & Quilt-V. & Quilt-V.-Red & Videopath & Bookset & Educontent & MMUpubmed \\
\midrule
\multicolumn{9}{c}{\textit{Training Strategy Ablations}} \\
\midrule
w/o data filter & 79.5 & 51.1 & 34.6 & 52.2 & 23.4 & 31.3 & 9.5 & 12.3 \\
w/o stain N/A & 77.7 & 52.1 & 35.1 & 57.3 & 28.4 & 32.0 & 11.1 & 12.3 \\
w/o curriculum & 76.3 & 46.7 & 30.9 & 52.6 & 18.4 & 29.3 & 10.5 & \textbf{13.1} \\
w/o native resolution & 79.5 & 46.3 & 29.0 & 46.2 & 23.8 & 25.5 & 7.3 & 10.6  \\
\midrule
\multicolumn{9}{c}{\textit{Inference Prompt Robustness}} \\
\midrule
No Prompt & 78.0 & 51.8 & 34.3 & 56.0 & 29.1 & 32.0 & 11.1 & 12.9 \\
No One Word & 78.5 & 51.5 & 34.8 & 56.3 & 30.0 & 33.8 & 11.5 & 12.6 \\
\midrule
HOMIE & \textbf{79.8} & \textbf{52.7} & \textbf{35.8} & \textbf{57.5} & \textbf{30.7} & \textbf{34.4} & \textbf{12.1} & \textbf{13.1} \\
\bottomrule
\end{tabular}
}
\label{t5}
\end{table}

\section{Conclusion}
In this paper, we formally define the task of Pathology Composed Retrieval (PCR) and introduce the first comprehensive benchmark to evaluate interleaved multi-modal queries. To overcome the Architectural mismatch of dual-encoders and the Task/Domain mismatches of generative models, we propose HOMIE—a model-agnostic, systematic adaptation framework. HOMIE transforms general Multimodal Large Language Models (MLLMs) into specialized pathology retrieval models. Ablations indicate that the two-stage design accounts for the observed performance. Stage 1 aligns the generative latent space for retrieval, while Stage 2 instills morphological priors through dynamic native-resolution processing, stain augmentation, and a progressive knowledge curriculum. Our lightweight 2B-parameter HOMIE variant substantially outperforms existing paradigms on composed retrieval, while maintaining strong performance on traditional tasks. By generating a single, unified multi-modal embedding for complex clinical queries, HOMIE provides a reliable retrieval system that empowers pathologists to find similar information and make their own independent, evidence-based diagnoses. Future work will extend this framework to integrate broader modalities, such as genomics and spatial transcriptomics, moving toward a clinical AI assistant for pathology.

\section*{Acknowledgements}
This work was partially supported by US National Science Foundation IIS-2412195, CCF-2400785, the Cancer Prevention and Research Institute of Texas (CPRIT) award (RP230363), the National Institutes of Health (NIH) R01 award (1R01AI190103-01) and Microsoft Accelerate Foundation Models Research (2024).

%
%
\bibliographystyle{splncs04}
\bibliography{main}
\clearpage
\appendix
\setcounter{page}{1}
    \begin{center}
    \LARGE
    \textbf{Supplementary Material}
     \\[20pt]
    \end{center}

\section{Clinical Application and Workflow}
\begin{figure}[t]
  \centering
\includegraphics[width=\linewidth]{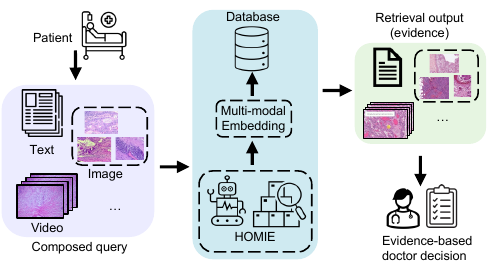}
\caption{\textbf{Clinical Application Workflow of HOMIE.} A complex clinical scenario involves a \textit{Composed query} containing interleaved \textit{Text}, \textit{Image}, and \textit{Video} inputs originating from a patient case. HOMIE processes this interleaved sequence into a single, unified \textit{Multi-modal Embedding} to search a massive historical Database. The \textit{Retrieval output (evidence)} is then presented to the pathologist, supporting an \textit{Evidence-based doctor decision} while fully preserving clinical interpretability and autonomy.}
   \label{teser}
\end{figure}

To better contextualize the practical utility of the proposed Pathology Composed Retrieval (PCR) task, Figure~\ref{teser} illustrates the end-to-end clinical workflow empowered by HOMIE. 

In a real-world diagnostic scenario, a pathologist rarely queries a system with a single, isolated image. Instead, the diagnostic process naturally yields a Composed query, an interleaved sequence of evidence that may include an initial H\&E patch (\textit{Image}), specific spatial or semantic instructions (\textit{Text}), or even dynamic panning sequences from a whole-slide scanner (\textit{Video}). 

Current AI paradigms face critical bottlenecks in this workflow. Dual-encoders fail to fuse these complex interleaved inputs, while generative MLLMs attempt to directly output a text report, introducing severe risks of clinical hallucinations and acting as opaque ``black boxes''.

HOMIE elegantly bridges this gap. By deep-fusing the entire interleaved sequence into a single, highly discriminative Multi-modal Embedding, HOMIE instantly queries a hospital's historical Database to find structurally and semantically identical reference cases. The resulting Retrieval output (evidence) serves as a powerful diagnostic anchor. Crucially, this retrieval-based paradigm preserves clinical autonomy: rather than blindly trusting an AI-generated report, the pathologist leverages the retrieved historical evidence to make their own \textbf{Evidence-based doctor decision}. This positions HOMIE as a highly reliable, interpretable, and safe AI assistant for precision computational pathology.

\section{Limitations and Future Work}

\noindent\textbf{Lack of Real-World Composed Queries.} A major challenge in this newly defined field is the absence of large-scale, clinically annotated datasets for composed retrieval. While we utilized GPT-5 to generate high-quality, heavily filtered composed query data, it might still exhibit slight distribution shifts from actual clinical workflows. \textit{Future Work:} We plan to collaborate with partner hospitals to collect real-world query logs from practicing pathologists. This will allow us to instruction-tune the model to perfectly align with the specific diagnostic habits and phrasing of human experts.

\noindent\textbf{Limitation of Data Modalities.} Currently, HOMIE focuses on deep-fusing visual (images, video) and textual data. However, modern comprehensive pathological diagnosis is increasingly multi-omics, integrating genomics, transcriptomics, and patient clinical history. \textit{Future Work:} A natural evolution of HOMIE is the integration of genomic and tabular encoders into the fusion backbone. We aim to extend the PCR task to handle queries such as ``Find historical cases with similar morphology and strictly matching gene mutation profiles,'' moving one step closer to a truly holistic diagnostic AI assistant.

\noindent\textbf{Computational Requirements and Deployment.} A common misconception regarding MLLM-based frameworks is that their massive scale prohibits practical clinical deployment. We explicitly clarify that while the \textit{training} phase of HOMIE requires substantial computational resources, the \textit{inference} phase is highly accessible and remarkably lightweight. Once trained, HOMIE can be easily deployed on standard, consumer-grade hardware. For instance, processing a standard 336px query requires only a 16GB GPU with a latency of approximately 50ms per query, while handling native high-resolution (1148px) inputs easily fits within a single 24GB GPU. \textit{Future Work:} To further optimize high-throughput retrieval for massive hospital-scale databases, we will explore pathology-specific model quantization (e.g., INT4/INT8) and KV-cache optimization, reducing the inference footprint without compromising the integrity of the semantic embedding space.

\section{Inference Pipeline}
Our model, trained primarily on retrieval tasks focused on image-text pairs, requires adaptation to handle zero-shot downstream tasks. The details of how we modify the model for these different zero-shot downstream tasks are introduced as follows. 

For the composed retrieval task, where the query consists of multiple images, or both an image and a question, or videos, we need to convert this multimodal information into a unified embedding. Thus, we use the prompt (i) for multiple images input, we use: \texttt{<image\textsubscript{1}><image\textsubscript{2}>...<image\textsubscript{i}> Summarize above images in one word: <emb>}; (ii) for mixed image-text input, we utilize: \texttt{<image\textsubscript{1}><text\textsubscript{1}>...<image\textsubscript{i}><text\textsubscript{j}> Summarize above image and sentence in one word: <emb>.} (iii) for video input, we use: \texttt{<video> Summarize above video in one word: <emb>}; where \texttt{<image>}, \texttt{<text>} and \texttt{<video>} denote placeholders for the input image, text and video, respectively. We use the last hidden state immediately preceding the \texttt{<emb>} token as the representation of the input.

For zero-shot classification, we adopt a prompt-based method inspired by the CLIP model~\cite{radford2021learning}. 
In this approach, each class name is expanded into a sentence using a specific template. We apply prompt ensembling to aggregate different prompts. Our model then computes embeddings for these sentences and the test images, extracting the last token of the MLLM output as described in Section 3.2. The similarity between these embeddings is calculated as outlined in Section 3.3, and test image labels are assigned based on the highest similarity scores.

\section{Training Data}
We leverage several large-scale public datasets to construct more image-text pairs for pathology-specific tuning.

The PathGen-1.6M Dataset contains over 1.6 million high-quality image-text pairs, making it currently one of the largest and most refined datasets in the pathology domain~\cite{sunpathgen}. The majority of this dataset is sourced from The Cancer Genome Atlas platform, a comprehensive, publicly funded project that provides clinical, genomic, and pathology data across various cancer types.

The PathCap Dataset contains approximately 223K image-caption pairs, among which 207K are high-quality pathology-specific examples~\cite{sun2024pathasst}. We select these high-quality pathology-specific examples.

The Quilt-1M Dataset primarily consists of approximately 600K pathology-related images and one million text descriptions, with many images linked to multiple captions~\cite{ikezogwo2024quilt}. Unlike traditional
academic sources, this dataset draws primarily from social media platforms. Therefore, we initially trained on QUILT-1M to obtain a baseline model and then filtered out low-similarity (similarity score $<$ 0.1) image–text pairs based on this model. Finally, we extracted around 500k image-text pairs.

\begin{figure}[!t]
  \centering
\includegraphics[width=\linewidth]{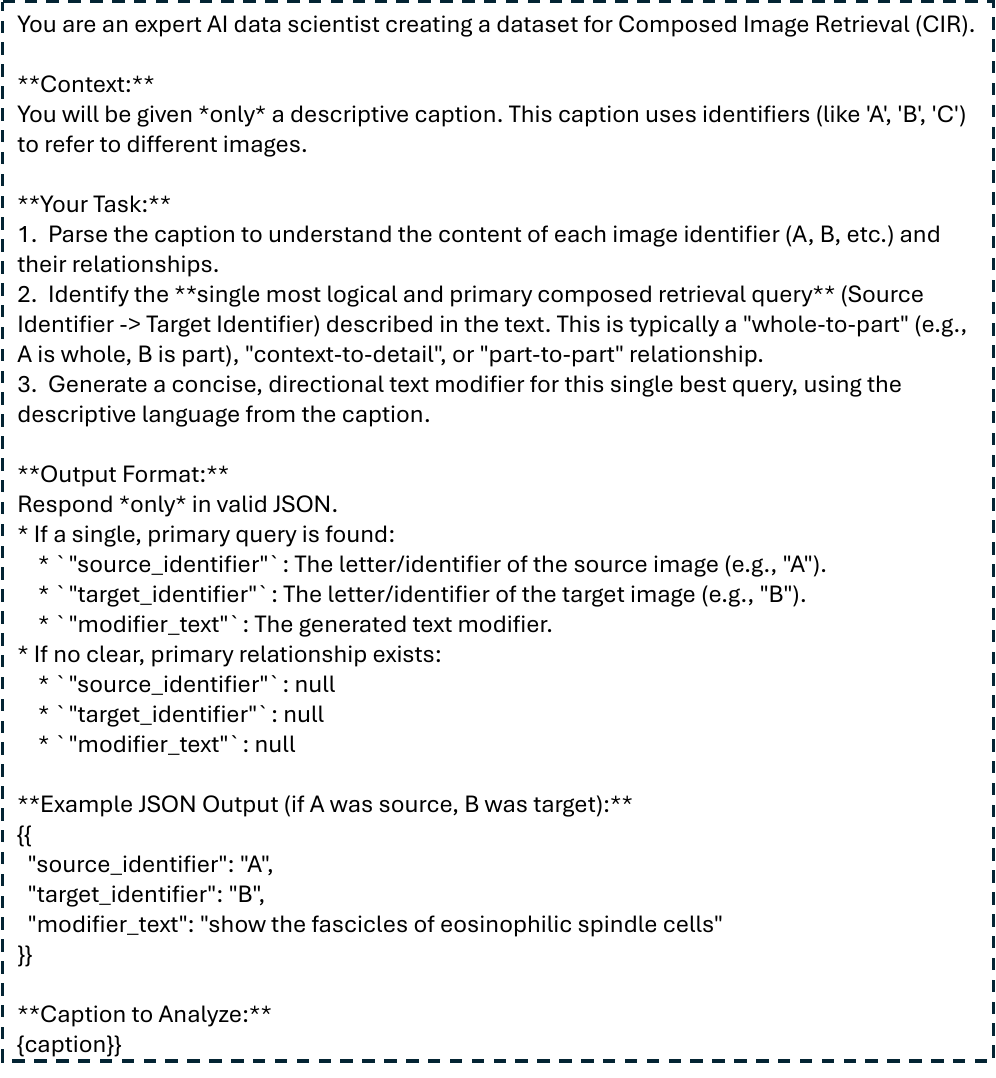}
\caption{Prompt to get Image-Text to Image Retrieval dataset}
   \label{s2}
\end{figure}

\section{Composed Retrieval Dataset Details}\label{7}
For Multi-Image to Text Retrieval, we extract multi-image subset with a single caption from Book-set~\cite{gamper2021multiple}, which is 600 samples in total.

For Image-Text to Image Retrieval, we use GPT-5 to analyze diagnostic reports from the multi-image set in Bookset~\cite{gamper2021multiple} and generate relational text. The prompt is shown in Fig~\ref{s2}. We extracted 488 samples.
\begin{figure}[!t]
  \centering
\includegraphics[width=\linewidth]{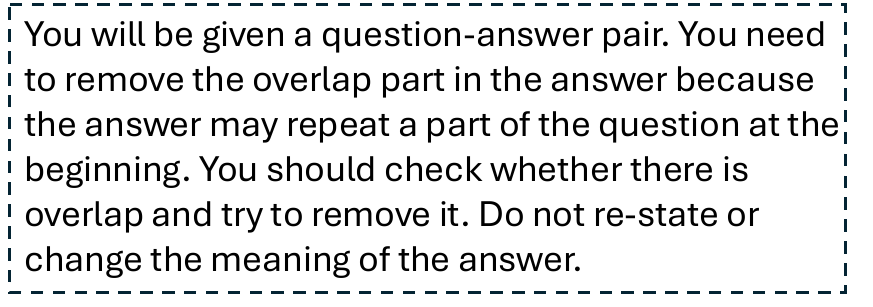}
\caption{Prompt to generate Image-Text to Text Retrieval dataset.}
   \label{s3}
\end{figure}

For Image-Text to Text Retrieval, we select only open-ended questions from the VQA data, where answers are more complex than a simple ``yes'' or ``no''. Additionally, since there is an overlap between questions and answers in the VQA data source, we utilize GPT5 for data cleaning to ensure clarity and accuracy in task evaluation. The prompt we use is shown in Fig~\ref{s3}. Finally, we get 724 samples from Quilt-VQA~\cite{ikezogwo2024quilt} and 252 samples from Quilt-VQA-RED~\cite{ikezogwo2024quilt}.

For Video to Text Retrieval, we use the Videopath~\cite{vuong2025videopath} dataset, which consists of educational histopathology videos and diagnostics. We extracted 244 samples from test data.

\end{document}